\documentclass[10pt,twocolumn,letterpaper]{article}

\usepackage{cvpr}              
\definecolor{cvprblue}{rgb}{0.21,0.49,0.74}
\usepackage[pagebackref,breaklinks,colorlinks,allcolors=cvprblue]{hyperref}

\usepackage{makecell}
\usepackage{multicol,multirow} 
\usepackage{threeparttable}

\definecolor{mygray}{gray}{.9}
\definecolor{mypink}{rgb}{.99,.91,.95}
\definecolor{mycyan}{cmyk}{.3,0,0,0}
\usepackage{colortbl}
\usepackage{graphicx}
\usepackage{float}
\usepackage{caption}
\usepackage{pifont} 

\def\confName{CVPR}
\def\confYear{2026}

\title{MagicWorld: Towards Long-Horizon Stability for Interactive \\ Video World Exploration}

\author{
  \vspace{-25pt}\\
  Guangyuan Li$^{1,2}$ ~~ 
  Bo Li$^2$ ~~
  Jinwei Chen$^2$ ~~
  Xiaobin Hu$^3$ ~~
  Lei Zhao$^{1,}$\textsuperscript{*} ~~
  Peng-Tao Jiang$^{2,}$\textsuperscript{*}
  \\
  $^1$College of Computer Science and Technology, Zhejiang University \\ $^2$vivo BlueImage Lab, vivo Mobile Communication Co., Ltd. \hspace*{1em} 
  $^3$National University of Singapore
  \\
  Project Page:~\, \url{vivocameraresearch.github.io/magicworld}
}

\begin{document}

\twocolumn[{%
\maketitle
\begin{center}
\includegraphics[width=\linewidth]{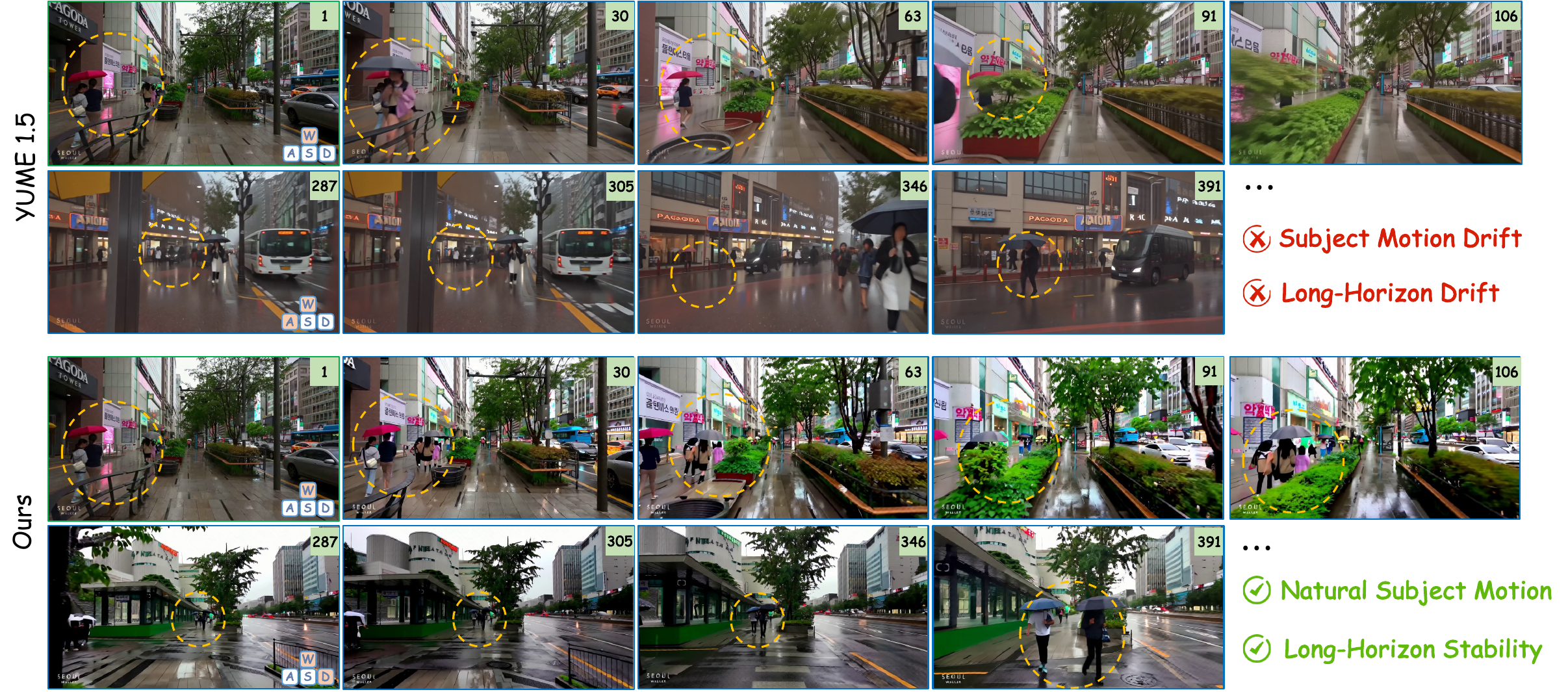}
\captionof{figure}{
We present MagicWorld, which addresses subject motion drift and long-horizon instability in video world models. Existing methods often exhibit unrealistic subject dynamics and unstable scene evolution as generation progresses. MagicWorld instead preserves natural subject motion and enables stable long-horizon generation, producing coherent dynamics and consistent scene structure over time. Yellow dashed circles highlight the subject in the scene, while the numbers in the top-right corner indicate the frame index.
}
\label{FIG_teaser}
\end{center}
}]

\renewcommand{\thefootnote}{\fnsymbol{footnote}}
\setcounter{footnote}{1}
\footnotetext{Corresponding author.}

\begingroup
\renewcommand\thefootnote{}      
\footnotetext{This work was completed by Guangyuan Li during internship at vivo. }
\addtocounter{footnote}{-1}      
\endgroup

\begin{abstract}
Recent interactive video world model methods generate scene evolution conditioned on user instructions. Although they achieve impressive results, two key limitations remain.
First, they exhibit motion drift in complex environments with multiple interacting subjects, where dynamic subjects fail to follow realistic motion patterns during scene evolution.
Second, they suffer from error accumulation in long-horizon interactions, where autoregressive generation gradually drifts from earlier scene states and causes structural and semantic inconsistencies.
In this paper, we propose MagicWorld, an interactive video world model built upon an autoregressive framework.
To address motion drift, we incorporate a flow-guided motion preservation constraint that mitigates motion degradation in dynamic subjects, encouraging realistic motion patterns and stable interactions during scene evolution.
To mitigate error accumulation in long-horizon interactions, we design two complementary strategies, including a history cache retrieval strategy and an enhanced interactive training strategy.
The former reinforces historical scene states by retrieving past generations during interaction, while the latter adopts multi-shot aggregated distillation with dual-reward weighting for interactive training, enhancing long-term stability and reducing error accumulation.
In addition, we construct RealWM120K, a real-world dataset with diverse city-walk videos and multimodal annotations to support dynamic perception and long-horizon world modeling.
Experimental results demonstrate that MagicWorld improves motion realism and alleviates error accumulation during long-horizon interactions. 
\end{abstract}

\section{Introduction}
\label{sec:intro}

Video world models \cite{agarwal2025cosmos,liu2024sora,parker2024genie,yang2023learning,yang2024position,feng2024matrix,guo2025mineworld} learn the evolution of visual scenes conditioned on actions, enabling agents to understand, predict, and plan in open environments. 
By capturing spatiotemporal structure, motion dynamics, and object interactions, they have shown strong potential in applications such as autonomous driving \cite{bar2025navigation,yang2025drivearena,liao2025diffusiondrive}, embodied intelligence \cite{xiao2025worldmem,liu2025embodied,ren2024embodied}, and virtual-world generation \cite{zhang2025matrix,agarwal2025cosmos,li2025hunyuan}. 
Furthermore, through generative video prediction and action-conditioned modeling, video world models \cite{feng2024matrix,guo2025mineworld,wu2025video,mao2025yume,zhang2025matrix,he2025matrix} can construct interactive virtual environments for low-cost experimentation and policy optimization, enabling stronger generalization and long-term planning.

Currently, interactive video world models \cite{wu2025video,mao2025yume,zhang2025matrix,he2025matrix,stableworld2026,lingbot-world} have become mainstream, and they are typically built using diffusion models combined with autoregressive generation. 
%
However, these methods still face two main challenges, as shown in Fig. \ref{FIG_teaser}.
(1) They suffer from motion drift in generated scenes, where dynamic subjects fail to follow realistic motion patterns during scene evolution. In real-world settings, interactions between dynamic subjects and static structures are highly coupled and nonlinear. Consequently, subjects that should undergo motion during scene transitions frequently remain static or exhibit degraded motion, undermining the realism of dynamic scene evolution.
(2) They lack effective strategies to mitigate error accumulation during long-horizon interactions. 
As generation proceeds autoregressively, small prediction errors gradually compound over time, leading to drift from earlier observations and causing structural and semantic inconsistencies in extended interactive scenarios.
%

Based on the above analysis, we propose MagicWorld, an interactive video world model.
Our method focuses on resolving motion drift in dynamic environments and alleviating error accumulation across extended interactive generation.
%
%
Specifically, we introduce a flow-guided motion preservation constraint to prevent motion drift and degradation of dynamic subjects. By enforcing temporal coherence in dynamically evolving regions, the constraint promotes realistic motion patterns and stable interactions with surrounding scene structures across frames.
To mitigate error accumulation during long-horizon interactive generation, we design a history cache retrieval strategy and an enhanced interactive training strategy.
%
The former preserves historical scene information across autoregressive steps by storing generated frame latents in a cache and retrieving recent historical representations. 
The retrieved representations are incorporated into subsequent generation steps, reinforcing previously established scene structures, maintaining view consistency across viewpoint changes, and reducing progressive drift.
%
The latter adopts an enhanced interactive training strategy based on multi-shot aggregated  distribution matching distillation (DMD) with dual-reward weighting. 
Specifically, unlike prior interactive DMD methods \cite{huang2025self,yang2025longlive} 
that update the generator after each interaction step, we aggregate the DMD losses across all interaction steps before performing optimization. 
This design allows the generator to evaluate the overall quality of the generated trajectory rather than optimizing each step independently. Furthermore, we introduce a dual-reward mechanism that weights the distillation objective using both visual quality and motion quality signals.
Finally, we construct the RealWM120K dataset, specifically designed to support dynamic object modeling in real-world scenes. It consists of city-walk videos collected from multiple cities worldwide and is accompanied by multimodal annotations, providing a comprehensive benchmark for studying dynamic perception and long-horizon world consistency.
Our contributions to the community are fourfold:
\begin{itemize}
    \item We propose MagicWorld, an autoregressive interactive video world model, designed to address motion drift in dynamic environments and mitigate error accumulation in long-horizon interactive generation.
    
    \item We introduce a flow-guided motion preservation constrain that enforces temporal coherence in dynamic regions to prevent motion drift and ensure realistic motion evolution of dynamic subjects.

    \item We design a history cache retrieval strategy to preserve historical scene states during autoregressive rollout, and an enhanced interactive training strategy based on multi-shot aggregated DMD with dual-reward weighting, jointly improving long-horizon stability and reducing error accumulation.

    \item We build the RealWM120K dataset with diverse city-walk videos and multimodal annotations for real-world video world modeling. Extensive experiments demonstrate that MagicWorld outperforms state-of-the-art methods in both VBench metrics and visual quality.
\end{itemize}

\section{Related Work}
\subsection{Video World Models}
Recently, video world models \cite{feng2024matrix,guo2025mineworld,wu2025video,mao2025yume,zhang2025matrix,he2025matrix,xiao2025worldmem,hyworld2025,yu2025cam,dai2025fantasyworld,stableworld2026,lingbot-world,worldplay2025,zheng2026versecrafter} have attracted increasing attention, with the goal of learning world dynamics directly from visual data and enabling controllable world evolution and state rollout.
For instance,
Matrix-Game 2 \cite{zhang2025matrix} introduced an interactive world model that enabled controllable game world generation and long-horizon video synthesis.
Yume \cite{mao2025yume} introduced a video world model that generated dynamically explorable worlds from a single input image and supported text-controlled world evolution.
Guo \textit{et al.} \cite{guo2025mineworld} proposed MineWorld, an interactive video world model designed for the Minecraft environment.
HY-World 1.5 \cite{hyworld2025} introduced a systematic interactive world modeling framework that maintained geometric consistency while satisfying real-time latency requirements.
Dai \textit{et al.} \cite{dai2025fantasyworld} proposed FantasyWorld, which jointly modeled video latent representations and implicit 3D fields within a single forward pass.
StableWorld \cite{stableworld2026} introduced long interactive video generation, with a focus on improving generation stability and temporal consistency.
LingBot-World \cite{lingbot-world} introduced an open-source world simulation framework based on video generation, supporting long-horizon video generation with low-latency real-time interaction.
However, existing video world models suffer from motion drift in dynamic subjects, where objects that should move remain static or exhibit degraded motion. To address these issues, we introduce a flow-guided motion preservation constraint to prevent motion degradation.

\subsection{Long Video Generation}
Current long video generation methods \cite{liarlon,lin2025autoregressive,xiang2025macro,yin2025slow,gu2025long,huang2025self,yang2025longlive,lu2025reward,yesiltepe2025infinity,huang2025memory,yu2025cam} typically combine diffusion models with autoregressive (AR) prediction, forming a middle-ground paradigm between purely diffusion-based and purely AR-based approaches.
For example,
CausVid \cite{yin2025slow} reconstructed video diffusion into a causal AR process and reduced inference steps through distribution-matching distillation.
MAGI-1 \cite{teng2025magi} scaled AR video generation via chunk-wise prediction.
Self-forcing \cite{huang2025self} narrowed the gap between training and inference in AR video diffusion by simulating inference during training.
LongLive \cite{yang2025longlive} performed causal frame-level AR, introducing KV-recache, long-sequence training to enhance long-term consistency and accelerate generation.
Reward Forcing \cite{lu2025reward} adopted reward-driven distribution matching, enabling real-time video generation with improved motion quality.

Existing methods \cite{yang2025longlive,xiao2025worldmem,yu2025cam} mitigate error accumulation in autoregressive generation by either retrieving historical information or adopting interactive training strategies.
For example, some approaches \cite{xiao2025worldmem,yu2025cam} retrieve historical frames based on camera trajectories and FOV overlap, while others maintain a memory bank of past frames and select relevant ones using a greedy matching strategy based on FOV overlap and temporal distance.
However, these methods rely heavily on geometric information and may miss semantically relevant historical states when viewpoints differ. 
%
In contrast, our method performs retrieval directly in the latent space using similarity matching, without requiring camera information.
This allows the model to retrieve historical states that are most relevant to the current generation in terms of structure or semantics, leading to more stable long-horizon and cross-interaction generation.
In addition, during interactive training, instead of updating parameters of the generator after each interaction step \cite{huang2025self,yang2025longlive}, we aggregate the DMD losses across all interaction steps before performing optimization. 
This allows the model to evaluate the overall quality of the generated video rather than optimizing each step independently, further improving long-horizon consistency and mitigating error accumulation.

\begin{figure*}[t]
\begin{center}
\includegraphics[width=\linewidth]{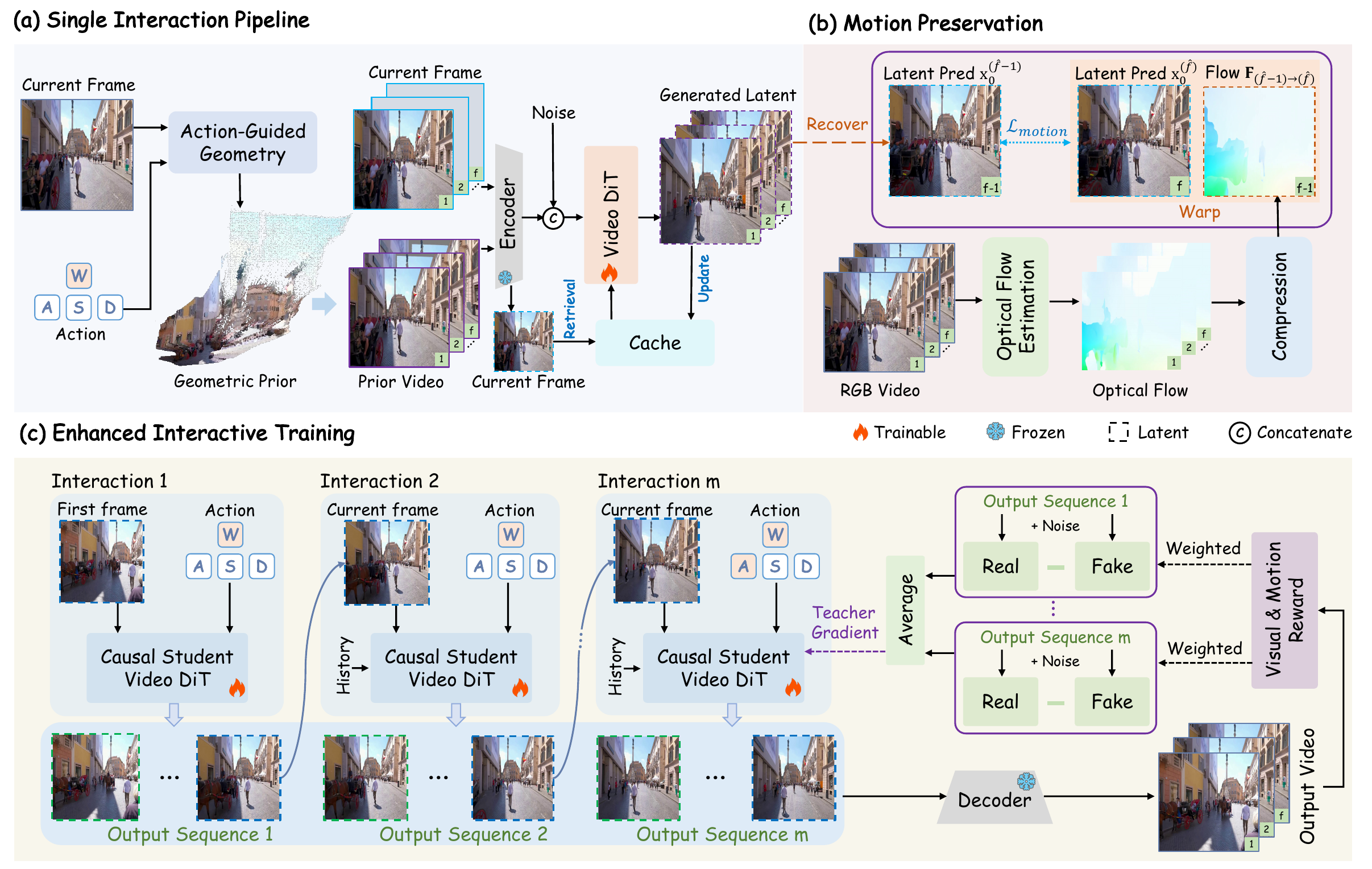}
\end{center}
\vspace{-15pt}
\caption{
Overview of MagicWorld.
(a) Single interaction pipeline with history cache retrieval.
(b) Flow-guided motion preservation enforces temporal coherence in dynamic regions.
(c) Enhanced interactive training with multi-shot aggregated reward DMD, jointly optimizing multi-step rollouts with visual and motion rewards to improve motion realism and mitigate error accumulation.
}
\label{FIG_Model}
\end{figure*}

\section{Methodology}
\subsection{Overview}
Our method aims to build an interactive video world model that allows users to explore a dynamically evolving world constructed from a single scene image through continuous interactions. 
We focus on addressing two key challenges in interactive world generation, the motion drift of dynamic subjects and instability in long-horizon interactions, enabling more realistic motion and stable scene evolution over extended rollouts.
The overall single interaction pipeline of MagicWorld is illustrated in Fig. \ref{FIG_Model}(a). Given a single scene image $\mathbf{I}_0 \in \mathbb{R}^{H \times W \times 3}$, as the initial observation for world construction, the model generates $f$ frames at each interaction step.
At the $n$-th step interaction, the user provides an action
$a_n \in \mathcal{A}$, where $\mathcal{A}$ denotes the action space, such as keyboard commands (\emph{e.g.}, combinations of W, A, S, and D). This action specifies the intended movement or viewpoint change within the virtual world.
Since generation at interaction step $n\!+\!1$ depends on the last frame produced at step $n$, we denote this frame as $\mathbf{I}_{n}^{(f)}$.
Thus, generation at step $n\!+\!1$ can be formulated as: 
$\mathbf{V}_{n+1} = G(\mathbf{I}_{n}^{(f)}, a_{n+1})$, 
where $\mathbf{V}_{n+1} = \{ \mathbf{I}_{n+1}^{(1)}, \mathbf{I}_{n+1}^{(2)}, \ldots, \mathbf{I}_{n+1}^{(f)} \}$ represents the $f$-frame video segment generated at step $n\!+\!1$.
$G(\cdot)$ denotes the video DiT.
After generation, the final frame $\mathbf{I}_{n+1}^{(f)}$ is used as the initial state for the next step to support subsequent interaction. 
Ultimately, under the action sequence $\{a_1, \ldots, a_{n}, \ldots\}$, MagicWorld produces the corresponding sequence of video segments $\{\mathbf{V}_1, \ldots, \mathbf{V}_{n}, \ldots\}$, thereby progressively constructing an explorable world.

\subsection{Motion and Geometry Preservation}
In video world generation, we observe that existing methods often suffer from motion drift during scene evolution. Dynamic subjects, such as pedestrians or vehicles, are expected to move consistently with scene transitions. However, in practice, they frequently remain static or exhibit incorrect motion patterns, resulting in motion collapse and unrealistic dynamics. This issue stems from the absence of explicit motion-aware constraints that couple dynamic behavior with scene evolution.
To address this limitation, we introduce a flow-guided motion preservation constraint that enforces temporal coherence in dynamically evolving regions, ensuring realistic motion patterns and stable interactions throughout interactive generation.
In addition, maintaining structural stability is equally critical. 
To preserve structural consistency, we incorporate an action-guided geometry strategy that grounds scene evolution in geometric constraints, thereby stabilizing spatial layouts during action-driven transitions.

\subsubsection{Flow-Guided Motion Preservation}
To prevent motion drift in dynamic subjects during scene evolution, we introduce a flow-guided motion preservation constraint that regularizes temporal coherence in motion-aware regions, as shown in Fig. \ref{FIG_Model}(b). 
%
During training, optical flow is estimated online from ground-truth videos using an optical flow estimation model \cite{teed2020raft}. To avoid the out-of-memory issue caused by optical-flow supervision in RGB space, motion supervision is performed in the latent space.
Following a velocity-based flow-matching formulation, we construct intermediate latent states via linear interpolation between the clean latent and noise: $\mathbf{x}_t=(1-t) \mathbf{x}_0+t \mathbf{x}_1$, where $\mathbf{x}_1 \sim \mathcal{N}(\mathbf{0}, \mathbf{I})$. The model predicts the velocity field $\mathbf{v}_\theta(\mathbf{x}_t, t)$, whose the target is defined as $\mathbf{x}_1 - \mathbf{x}_0$. Given the predicted velocity, we reconstruct the clean latent as: 
\begin{equation}
\mathbf{x}_0^{\text {pred }}=\mathbf{x}_t-t \mathbf{v}_\theta\left(\mathbf{x}_t, t\right).
\end{equation}
%
Motion supervision is applied on $\mathbf{x}_0^{\text {pred}}$, ensuring that the motion regularization targets the denoised content rather than the noisy intermediate.
Since motion supervision is applied in the latent space, the optical flow $\mathbf{F}$ is compressed to align with $\mathbf{x}_0^{\text{pred}}$ in dimension.
Let $\mathbf{x}_0^{(\hat{f}-1)}$ and $\mathbf{x}_0^{(\hat{f})}$ denote two consecutive latent frames from $\mathbf{x}_0^{\text {pred}}$, and let $\mathbf{F}_{(\hat{f}-1) \rightarrow (\hat{f})}$ be the forward flow. 
Here, $\hat{f} = f//4$ denotes the number of frames in the latent space, $4$ denotes the temporal compression ratio.
We warp $\mathbf{x}_0^{(\hat{f})}$ back to latent frame $\hat{f}-1$ using warping operator $\mathcal{W}(\cdot)$:
\begin{equation}
\tilde{\mathbf{x}}_0^{(\hat{f}-1)}=\mathcal{W}\left(\mathbf{x}_0^{(\hat{f})}, \mathbf{F}_{(\hat{f}-1) \rightarrow (\hat{f})}\right).
\end{equation}
This warping aligns the content of latent frame $\hat{f}$ to the coordinate system of latent frame $\hat{f}-1$, enabling direct temporal consistency measurement.
We define a motion-aware weighting based on the magnitude of the optical flow: $\mathbf{w}_{(\hat{f}-1)}=\left\|\mathbf{F}_{(\hat{f}-1) \rightarrow (\hat{f})}\right\|_2$, where 
the L2 norm measures the flow magnitude and assigns larger weights to regions undergoing stronger motion.
The flow-guided motion preservation constraint is defined as: 
\begin{equation}
\mathcal{L}_{\text {motion }}=\frac{1}{N} \sum_{\hat{f}} \sum_i \mathbf{w}_{(\hat{f}-1)}(i)\left|\mathbf{x}_0^{(\hat{f}-1)}(i)-\tilde{\mathbf{x}}_0^{(\hat{f}-1)}(i)\right|,
\end{equation}
where $i$ denotes the spatial location index, and $N$ represents the total number of elements involved in the summation. This formulation encourages temporal consistency in motion-aware regions while avoiding excessive constraints on static background areas. By explicitly regularizing regions with significant motion, the model mitigates motion drift and degraded motion behaviors, ensuring that dynamic subjects maintain coherent movement during scene evolution. Finally, the overall training objective combines the velocity flow-matching loss and the motion preservation loss:
$\mathcal{L}_{\text{stage1}} = \mathcal{L}_{FM} + \mathcal{L}_{\text {motion}}$.

\subsubsection{Action-Guided Geometry} 
To preserve structural consistency during action-driven scene evolution, we employ an action-guided geometry (AGG) strategy. Specifically, we reconstruct a 3D scene representation from the first frame and transform it according to action-induced viewpoint changes. The reconstructed geometry is projected into multiple views to encode action-conditioned scene transformations into a unified geometric prior. This prior serves as an explicit structural constraint for the world model, guiding scene evolution and stabilizing spatial layouts across interaction steps. AGG consists of two steps. 

\noindent \textit{\textbf{Action mapping.}}
To convert user interactions into camera trajectories, each discrete action 
$a_n \in \mathcal{A}$ is mapped to a camera extrinsic sequence of length $f$ (i.e., the trajectory at step $n$):
\begin{equation}
\begin{split}
\mathcal{T}\left(a_n ; \Theta\right) &= 
\left\{ \left(\mathbf{R}_n^{(k)}, \mathbf{t}_n^{(k)} \right) \right\}_{k=1}^f, \\
\left(\mathbf{R}_n^{(0)}, \mathbf{t}_n^{(0)}\right) &\equiv 
\left(\mathbf{R}_{n-1}^{(f)}, \mathbf{t}_{n-1}^{(f)}\right),
\end{split}
\end{equation}
where $\Theta$ denotes tunable parameters (e.g., step size, rotation angle, interpolation steps), $k = 1,\dots,f$ indexes the discretized camera poses along the trajectory, and $\mathbf{R}, \mathbf{t}$ are rotation and translation matrices.
Given the first frame $\mathbf{I}_n^{(1)}$, we estimate its depth $D(x)$ \cite{Bochkovskii2024} and back-project pixels using camera intrinsics $\mathbf{K}$ to obtain a 3D geometric prior:
$\hat{x}=\mathbf{K}^{-1}x,\; X_c=D(x)\hat{x}$.
The representation is then transformed into the world coordinate system to form the scene prior $\mathbf{P}$.

\noindent \textit{\textbf{Geometry projection.}}
At step $n$, the user provides an action $a_{n} \in \mathcal{A}$, which is mapped by the action controller to a corresponding camera pose $(\mathbf{R}_{n}^{(k)}, \mathbf{t}_{n}^{(k)})$.
Based on this camera pose, the geometric prior $\mathbf{P}$ in the world coordinate system is projected into the new viewpoint, producing the action-guided geometric representation:
\begin{equation}
\mathbf{P}^{action,(k)}_{n}  = \Pi(\mathbf{P}, \mathbf{K}, \mathbf{R}_{n}^{(k)}, \mathbf{t}_{n}^{(k)}),
\end{equation}
where $\Pi(\cdot)$ denotes the geometric projection operator.
The resulting $\mathbf{P}^{action,(k)}_{n}$ serves as an explicit geometric prior. 
We render the action-guided geometric sequence into a geometric prior video:
$\mathbf{V}^{prior}_{n} = \mathcal{R}\Big( \big\{ \mathbf{P}^{\text{action},(k)}_{n} \big\}_{k=1}^{f} \Big)$,
where $\mathcal{R}(\cdot)$ denotes the rendering function. 
The prior video $\mathbf{V}^{prior}_{n}$ is combined with the current first frame and the noise input, and fed into the Video DiT.

\subsection{Long-Horizon Interactive Learning}
Interactive video world generation follows an autoregressive process in which each step depends on previously generated results. As interactions continue, prediction errors accumulate and gradually cause scene drift, leading to structural instability and semantic deviations in long-horizon scenarios.
To address this issues, we introduce two complementary strategies, a history cache retrieval strategy to reinforce historical scene information during rollout and an enhanced interactive training strategy that explicitly simulates multi-step interaction during training to enhance long-term consistency.

\subsubsection{History Cache Retrieval}
The goal of history cache retrieval is to store the clean latent features generated at each autoregressive step into a history cache. During the subsequent inference step, the current input frame latent is matched against the cached history via similarity retrieval to obtain the most relevant reference frames, which are then fed into the model as auxiliary conditions. History cache retrieval consists of three steps.

\noindent \textit{\textbf{Cache update.}} 
Given the latent sequence 
$\{\mathbf{L}_n^{(1)}, \ldots, \mathbf{L}_n^{(\hat{f})}\}$ 
generated at step $n$.
We retain only its last-frame latent $\mathbf{L}_n^{(\hat{f})}$ for the next interaction, while the remaining frame latents are appended to the cache pool. 
We set the cache capacity to 20 latent frames. 
When updating the history cache, we keep the first-frame latent fixed and only update the remaining entries. 
The first latent is obtained from the initial interaction and directly corresponds to the latent derived from the input scene image, thus carrying the most essential and stable information about the environment. 
Once the cache reaches its capacity, the non-fixed entries are replaced in a first-in, first-out manner.
 
\noindent \textit{\textbf{Cache retrieval.}} 
At interaction step $n$, we take the latent of the first frame in the current step as $\mathbf{q}_{n}$. To perform retrieval, we apply spatial pooling to both the query and all cached latents to obtain vector representations:
$\mathbf{q} = \mathrm{pool}(\mathbf{q}_{n}), \mathbf{c}_i = \mathrm{pool}(\mathbf{h}_i)$.
Following previous works \cite{lu2021masa,li2022transformer}, we compute the cosine similarity:
\begin{equation}
    s_i = \frac{\langle \mathbf{q}, \mathbf{c}_i \rangle}{\|\mathbf{q}\| \, \|\mathbf{c}_i\|}.
\end{equation}
All cached latents are ranked by similarity, and the top-3 entries are selected.
Notably, retrieval is independent of temporal distance.
By computing similarity in the latent space, the model retrieves historical states that best match the current iteration in structure or semantics, even if they are from much earlier steps.
This design avoids bias toward short-term neighbors and provides the model with relevant contextual cues that help mitigate scene drift.

\noindent \textit{\textbf{Cache injection.}} 
The retrieved top-3 latents 
are embedded as history tokens and concatenated with the input tokens along the sequence dimension, providing explicit reference information. History cache retrieval mitigates autoregressive error accumulation by caching and retrieving historical information, supporting stable scene evolution and improved viewpoint consistency across interactions.

\subsubsection{Enhanced Interactive Training Strategy}
To mitigate error accumulation and enhance long-horizon stability, we adopt an enhanced interactive training strategy, as shown in Fig. \ref{FIG_Model}(c).
The overall optimization procedure consists of two stages, ordinary differential equation (ODE) initialization and multi-shot aggregated reward distribution matching distillation (DMD) training.

\noindent \textbf{\textit{ODE initialization.}} We start from a pretrained bidirectional video DiT trained with the $\mathcal{L}_{\text{stage1}}$ objective and reformulate it as a deterministic generative process via ODE initialization. This conversion enables causal generation while preserving the distribution learned by the original diffusion model.
Based on the ODE-initialized model, we construct a causal student generator $\hat{G}(\cdot)$.

\noindent \textbf{\textit{Multi-shot aggregated distillation.}}
During training, we explicitly simulate interactive rollout for $M$ steps. At interaction step $m$, given the current state $\mathbf{L}_m$ and action $a_m$, the student generates a latent segment: $\hat{\mathbf{X}}_m = \hat{G}(\mathbf{L}_m, a_m)$, and updates the state via $\mathbf{L}_{m+1} = \hat{\mathbf{X}}_m^{(f)}$. This process exposes the model to its own predictions and aligns with the interactive inference procedure. For each step, we compute a DMD loss $\mathcal{L}_{\text{DMD}}^{(m)}$.
However, unlike prior DMD-based approaches \cite{huang2025self,yang2025longlive} that update the generator immediately, we aggregate the distillation losses over the entire $M$-step interaction rollout before performing optimization:
\begin{equation}
\mathcal{L}_{\mathrm{DMD}}^{\mathrm{agg}}=\frac{1}{M} \sum_{m=1}^M \mathcal{L}_{\mathrm{DMD}}^{(m)}.
\end{equation}
This aggregated optimization allows the generator to perceive the overall quality of the trajectory rather than optimizing each step independently. By delaying parameter update until the full rollout is completed, the generator can evaluate later interaction outcomes and adjust earlier predictions accordingly, improving long-horizon consistency and mitigating error accumulation.

\noindent \textbf{\textit{Dual-reward quality guidance.}}
To further enhance long-horizon stability, we introduce reward weighting at the interaction level. For each generated segment $\hat{\mathbf{X}}_m$, we decode it into RGB and evaluate it using a pretrained VideoReward model \cite{liu2025improving}, which produces visual quality score $r_m^{\text{VQ}}$ and motion quality score $r_m^{\text{MQ}}$.
These scores assess visual consistency and motion fidelity in multi-step interactive video generation. We combine them as: $r_m=r_m^{\text{VQ}}+r_m^{\text{MQ}}$.
The reward-weighted multi-shot aggregated DMD objective becomes: 
\begin{equation}
\mathcal{L}_{\text {stage2 }}=\frac{1}{M} \sum_{m=1}^M \exp \left(r_m\right) \mathcal{L}_{\mathrm{DMD}}^{(m)}.
\end{equation}
By jointly optimizing over full interaction rollouts and weighting distillation with both visual and motion quality signals, the proposed strategy aligns training with interactive inference and enhances long-horizon stability.
Unlike prior distillation approaches \cite{huang2025self,yang2025longlive,lu2025reward}, our enhanced interactive training strategy aggregates multi-step supervision before parameter updates, enabling the model to perceive and optimize overall interaction quality across extended sequences.

\begin{figure*}[t]
\begin{center}
\includegraphics[width=\linewidth]{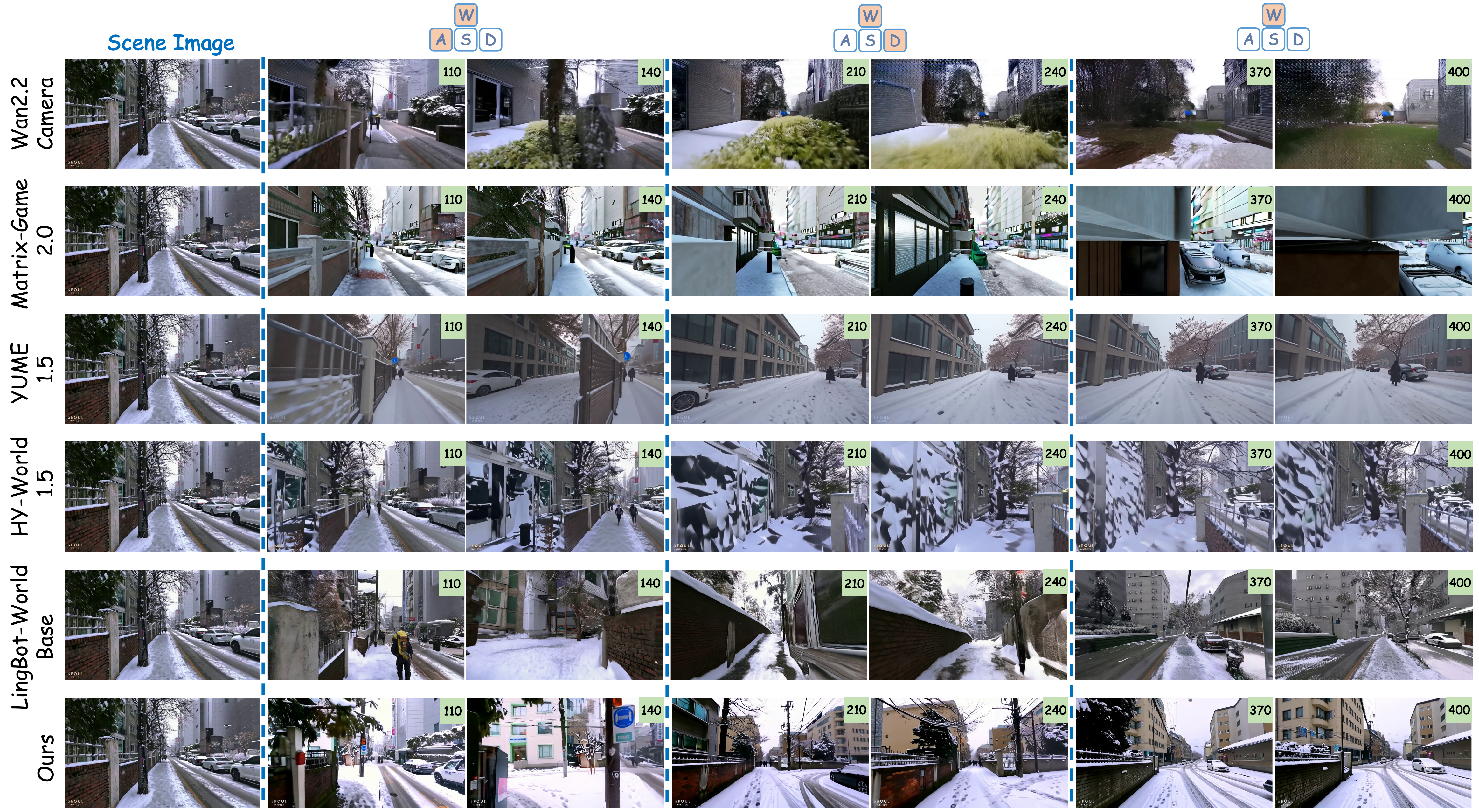}
\end{center}
\vspace{-15pt}
\caption{
Qualitative comparison of different methods on the same scene image under long-term interactions. Our method maintains stable scene geometry and natural motion dynamics while significantly reducing error accumulation, leading to more consistent generation during long-horizon interactions. Please \textbf{zoom in} for details. 
}
\label{Fig_res_main}
\end{figure*}

\subsection{Dataset Construction}
Recent advances in video world models \cite{feng2024matrix,guo2025mineworld,zhang2025matrix,he2025matrix,xiao2025worldmem,stableworld2026} have demonstrated strong performance in game environments and relatively static or weakly dynamic scenes. However, these models still struggle in real-world settings, especially in street scenes involving dense dynamic agents and non-trivial camera motion. 
To bridge this critical gap, we construct RealWM120K, a video dataset specifically designed for real-world environments, with a focus on city-walk street scenes collected across the globe. The dataset covers major cities from multiple countries and regions, spanning diverse conditions across different times of day and seasons, thereby providing a rich and realistic distribution of real-world city street scenarios. 
The dataset construction consists of four steps:

\noindent \textit{\textbf{Video collection.}} We collect city-walk videos from YouTube, which naturally capture unconstrained camera motion, diverse city layouts, and real-world dynamic agents. A subset of the video URLs is obtained from the Sekai \cite{li2025sekai} collection. In total, we collect 2,659 videos with over 4,000 hours of raw footage.

\noindent \textit{\textbf{Clip extraction.}} Raw videos are segmented into 140K short clips, each with a duration of 10 seconds at 30 FPS and a resolution of 720$\times$1280, balancing temporal continuity and computational feasibility for world model training. 

\noindent \textit{\textbf{Clip filtering.}} To ensure semantic coherence and visual quality, we employ Qwen2.5-VL-72B \cite{Qwen2VL} to assess both semantic quality and aesthetic quality for each clip. Starting from approximately 140K candidate clips, we retain the top 85$\%$, resulting in about 120K video clips used in the final dataset. 

\noindent \textit{\textbf{Multi-modal annotation.}} 
Each clip is annotated with a caption generated by Qwen2.5-VL-72B \cite{Qwen2VL}, focusing on scene layout, spatial structure, dynamic objects, lighting, weather, and overall atmosphere, enabling semantic grounding for world models. 
We extract synchronized geometric representations, including camera trajectories, point clouds, and object masks extracted using ViPE \cite{huang2025vipe}, and depth maps estimated by Video Depth Anything \cite{video_depth_anything}.
These modalities provide explicit supervision for spatial consistency, geometry-aware reasoning, and dynamic object understanding in real-world environments.

\begin{figure*}[t]
\begin{center}
\includegraphics[width=\linewidth]{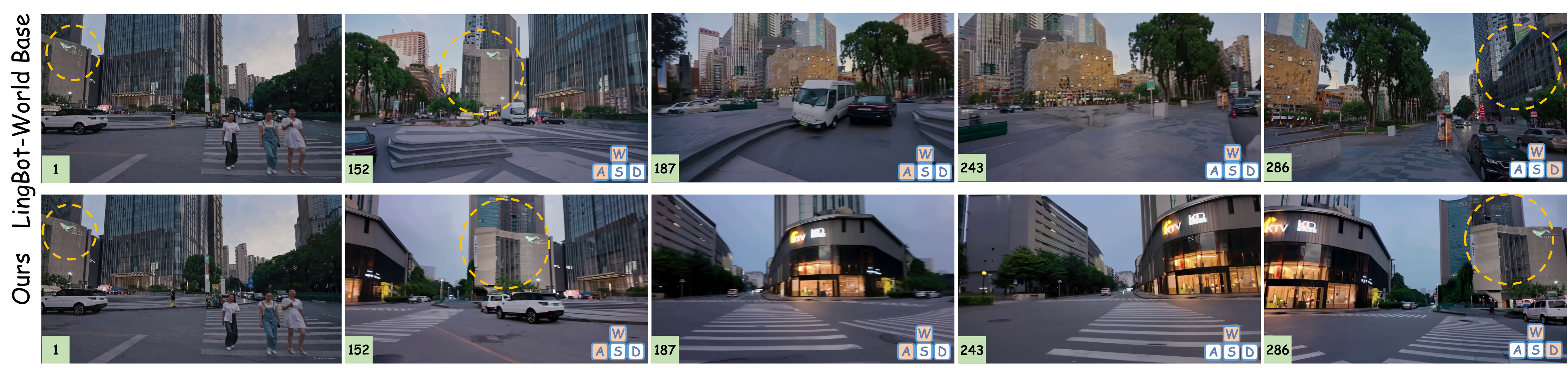}
\end{center}
\vspace{-15pt}
\caption{
Visual comparison of viewpoint consistency. Our method maintains consistent scene appearance across viewpoint changes, ensuring the scene remains consistent when revisiting the same viewpoint after multiple transitions. Please \textbf{zoom in} for details. 
}
\label{Fig_res_sub}
\end{figure*}
\begin{figure*}[t]
\begin{center}
\includegraphics[width=\linewidth]{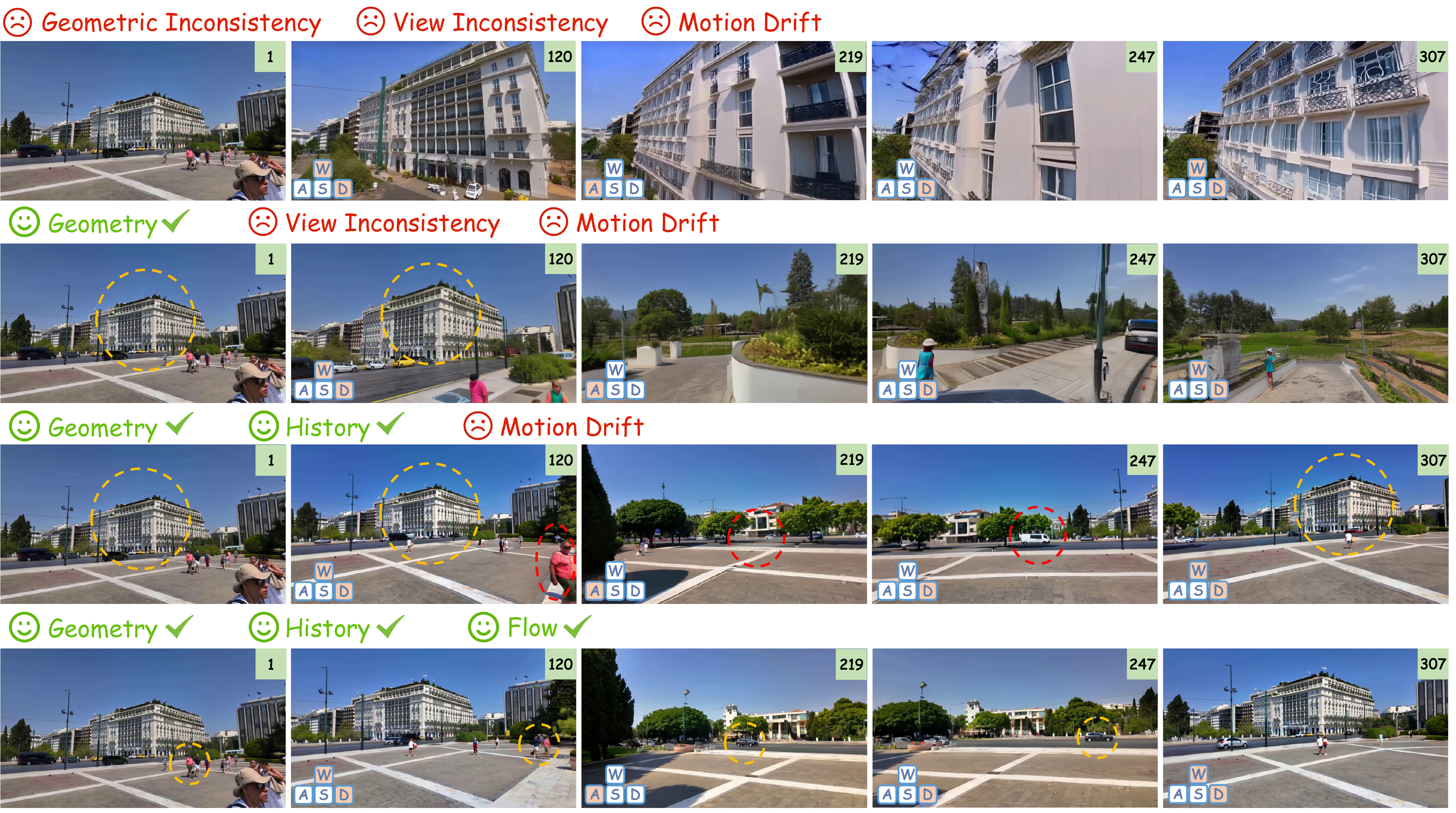}
\end{center}
\vspace{-15pt}
\caption{
Qualitative comparison of different component variants in the ablation study. The first row shows results from the bare model. As each component is progressively incorporated, the generated results show clear improvements in scene consistency, viewpoint consistency, and subject motion quality. Please \textbf{zoom in} for details. 
}
\label{Fig_ab_1}
\end{figure*}

\section{Experiments}

\subsection{Implementation Details}
We adopt the pretrained weights from Wan2.1-Fun-V1.1-1.3B \cite{WanFun1_3B} as the foundational model.
All training experiments are conducted on 56 NVIDIA A800 GPUs.
For bidirectional video DiT training, the batch size is set to 4 per GPU. The input video resolution is 480$\times$832, and each training sample contains 81 frames. We train the bidirectional model for 60k steps with a learning rate of 2$\times$10$^{-5}$.
For distillation training, we first collect 40k ODE pairs and fine-tune the causal student model for 10k steps. This is followed by an additional 10k steps of DMD-based training. The learning rate is set to 6$\times$10$^{-6}$. The multi-shot DMD rollout length is 4. The chunk size of latent frames, the local attention window size, and the sink size are set to 3, 9, and 3, respectively.
We evaluate on RealWM120K-Val, which contains 100 real-world images covering diverse scenes, generating videos at 480$\times$832 resolution with 81 frames at 16 FPS during inference.

\begin{table*}[t]
\centering
\setlength{\tabcolsep}{0.7mm}
\caption{Comparison of interactive video world models and camera-based video generation methods under VBench \cite{huang2024vbench} on the RealWM120K-Val dataset. The best and second-best results are highlighted in \textcolor{red}{red} and \textcolor{blue}{blue}, respectively. Latency (s) is measured with a batch size of 1 on a $480\times832\times81$ video using each model’s default inference steps for a fair comparison, and is evaluated on the NVIDIA H20 GPU.}
\vspace{-10pt}
\begin{threeparttable}
\resizebox{\linewidth}{!}{
\begin{tabular}{l|*{11}{c}}
\toprule
\multirow{2}{*}{\makecell{\\ Methods}}
& \multicolumn{2}{c}{Efficiency}
&& \multicolumn{4}{c}{Temporal Quality} 
&& \multicolumn{2}{c}{Visual Quality} 
 \\
\cmidrule{2-3}\cmidrule{5-8}\cmidrule{10-11}
& \makecell{Latency(s)}
 & \makecell{Overall\\Score $\uparrow$} 
 && \makecell{Temporal\\Flick. $\uparrow$} 
 & \makecell{Motion\\smooth. $\uparrow$}
 & \makecell{Subject\\Cons. $\uparrow$}
 & \makecell{Background\\Cons. $\uparrow$}
 && \makecell{Aesthetic\\Qua. $\uparrow$}
 & \makecell{Image\\Qua. $\uparrow$}
 
 \\
\midrule

ViewCrafter \cite{yu2024viewcrafter} 
& 302 & 0.7807
&& 0.9569 & 0.9790 & 0.8188 & 0.8748 
&& 0.5001 & 0.5543 
 \\

Wan2.1-Camera \cite{WanFun1_3B} 
& 39 & 0.8172 
&& 0.9586 & 0.9801 & 0.8778 & 0.9173
&& 0.5018 & 0.6674 
\\

Wan2.2-Camera \cite{WanFun5B} 
& 94 & 0.7935
&& 0.9573 & 0.9846 & 0.8508 & 0.8982
&& 0.4861 & 0.5837 
 \\

Matrix-Game 2.0 \cite{he2025matrix} 
& \cellcolor{mypink} \textcolor{red}{11} & 0.8082
&&  0.9457 & 0.9814 & 0.8476 & 0.8990
&& 0.4971 & 0.6784 
 \\

YUME \cite{mao2025yume} 
& 1918 & 0.8314 
&& 0.9491 & 0.9865 
& 0.9098
& 0.9264
&& 0.5239 
& 0.6926 
\\

YUME 1.5 \cite{mao2025yume1_5} 
& 19 & 0.8334
&& 0.9502 & 0.9821 
& 0.9115 
& 0.9221
&& 0.5270 
& \cellcolor{mycyan} \textcolor{blue}{0.7077} 
 \\

HY-World 1.5 \cite{hyworld2025} 
& 163 & \cellcolor{mycyan} \textcolor{blue}{0.8452}
&& \cellcolor{mycyan} \textcolor{blue}{0.9719} & \cellcolor{mycyan} \textcolor{blue}{0.9918} 
& 0.9409 
& \cellcolor{mycyan} \textcolor{blue}{0.9348}
&& 0.5303 
& 0.7014 
 \\

LingBot-World-Base \cite{lingbot-world} 
& 1920 & 0.8364
&& 0.9653 & 0.9755 
& \cellcolor{mycyan} \textcolor{blue}{0.9423} 
& 0.9261
&& \cellcolor{mypink}\textcolor{red}{0.5406} 
& 0.6687 
 \\

Ours 
& \cellcolor{mycyan} \textcolor{blue}{15} 
& \cellcolor{mypink} \textcolor{red}{0.8547}
&\cellcolor{mypink}& \cellcolor{mypink} \textcolor{red}{0.9752} & \cellcolor{mypink} \textcolor{red}{0.9921} & \cellcolor{mypink}\textcolor{red}{0.9627} & \cellcolor{mypink}\textcolor{red}{0.9408} &\cellcolor{mycyan}& \cellcolor{mycyan}\textcolor{blue}{0.5394} & \cellcolor{mypink}\textcolor{red}{0.7182}
 \\

\bottomrule
\end{tabular}
}
\end{threeparttable}
\label{tab_main}
\end{table*}

\begin{figure*}[t]
\begin{center}
\includegraphics[width=\linewidth]{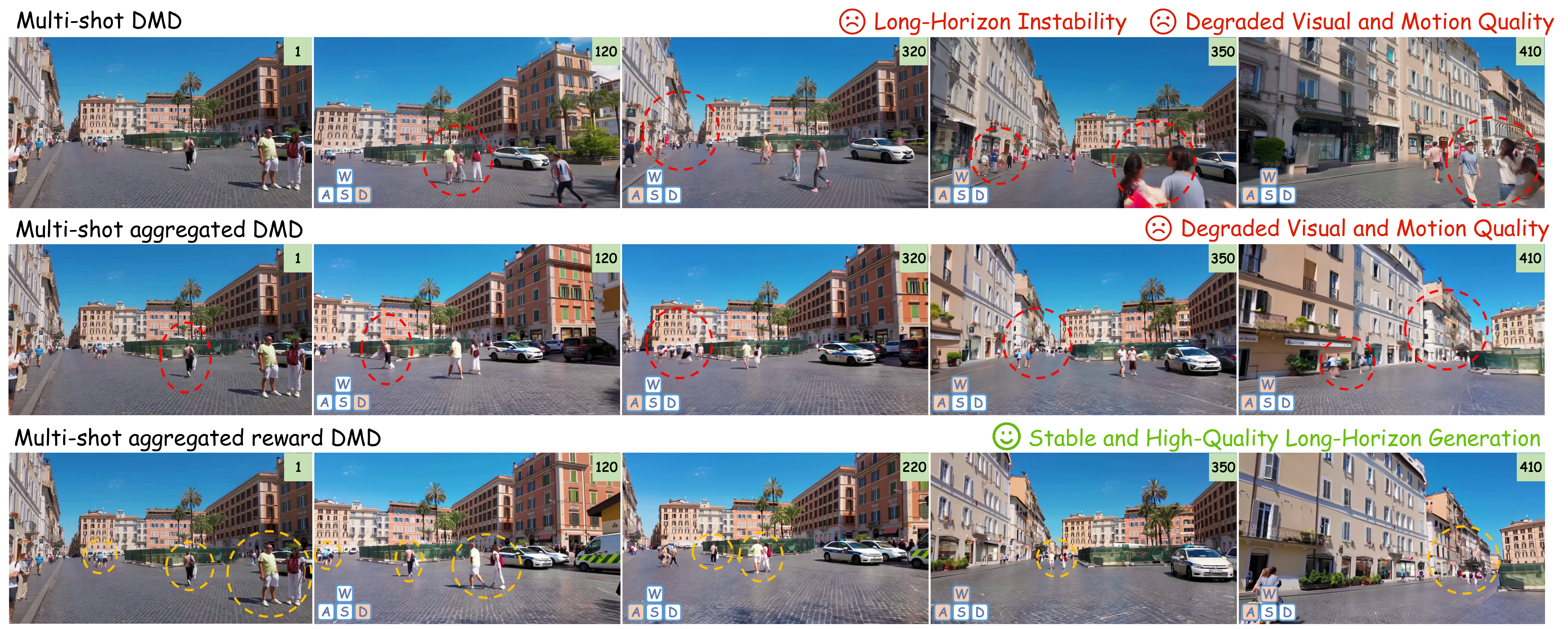}
\end{center}
\vspace{-15pt}
\caption{
Visual comparison of different DMD training schemes. Multi-shot aggregated reward DMD produces more stable and higher-quality long-horizon video generation. Red dashed circles indicate low-quality regions, while yellow dashed circles highlight high-quality regions. Please \textbf{zoom in} for details. 
}
\label{Fig_ab_2}
\end{figure*}

\subsection{Comparison with SOTA Methods}
We compare our method with recent interactive video world models as well as camera-based video generation models.
For a fair comparison, we adopt the autoregressive inference strategy for all camera trajectory methods, where the last frame of each interaction serves as the first frame of the subsequent one.
We use VBench \cite{huang2024vbench} metrics for evaluation, covering temporal and visual quality.

\noindent \textit{\textbf{Qualitative Comparison.}}
Fig. \ref{Fig_res_main} presents qualitative comparisons of long video sequences generated through continuous interactions within the same scene. Compared with existing methods, our approach demonstrates superior geometric preservation and motion consistency, while exhibiting significantly reduced error accumulation over extended interactions.
We further evaluate viewpoint consistency, as shown in Fig. \ref{Fig_res_sub}. Our method maintains strong scene consistency during viewpoint changes. Specifically, when the camera leaves a particular viewpoint and returns after multiple interaction steps, the original scene structure and dynamic subjects are still well preserved.
\textit{Additional video comparisons are provided in the supplementary material.}

\noindent \textit{\textbf{Quantitative  Comparison.}}
Tab. \ref{tab_main} reports the quantitative comparison on the RealWM120K-Val benchmark using VBench metrics.
Our method achieves the best overall score (0.8547), outperforming previous methods across efficiency, temporal quality, and visual quality.
In terms of efficiency, our method achieves competitive latency (15s), ranking second while being significantly faster than most existing world models.
For temporal quality, our method obtains the best results across all metrics, demonstrating strong temporal stability and dynamic preservation during interactive generation.
For visual quality, our method ranks second in aesthetic quality and achieves the best image quality score, indicating that the proposed geometry and motion preservation mechanisms improve temporal coherence without sacrificing visual fidelity.

\begin{table*}[t]
\centering
\setlength{\tabcolsep}{0.7mm}
\caption{Quantitative comparison of different variants using the RealWM120K-Val dataset. Overall score denotes the average over all VBench \cite{huang2024vbench} metrics.}
\vspace{-10pt}
\begin{threeparttable}
\resizebox{\linewidth}{!}{
\begin{tabular}{l|c|*{9}{c}}
\toprule
\multirow{2}{*}{\makecell{\\ Methods}} & \multirow{2}{*}{\makecell{\\ Overall Score}} & \multicolumn{4}{c}{Temporal Quality} && \multicolumn{2}{c}{Visual Quality} &  \\
\cmidrule{3-6}\cmidrule{8-10}
 & &  \makecell{Temporal\\Flick. $\uparrow$} 
 & \makecell{Motion\\smooth. $\uparrow$}
 & \makecell{Subject\\Cons. $\uparrow$}
 & \makecell{Background\\Cons. $\uparrow$}
 && \makecell{Aesthetic\\Qua. $\uparrow$}
 & \makecell{Image\\Qua. $\uparrow$}    \\
\midrule
Base Model & 0.8238 & 0.9599 & 0.9807 & 0.9075 & 0.9182 && 0.5023 & 0.6742 &  \\
\hspace{0.6em} Geometry \ding{51} & 0.8391 & 0.9690 & 0.9897 & 0.9361 & 0.9288 && 0.5174 & 0.6936 & \\
\hspace{0.6em} Geometry \ding{51} History \ding{51}  & 0.8412 & 0.9701 & 0.9901 & 0.9373  & 0.9294 && 0.5258  & 0.6945 & \\
\hspace{0.6em} Geometry \ding{51} History \ding{51} Flow \ding{51}  & 0.8471 & 0.9709 & 0.9912 & 0.9599 & 0.9334  && 0.5278 & 0.6995   & \\

\midrule
Multi-shot DMD & 0.8463 & 0.9717 & 0.9889 & 0.9585 & 0.9326 && 0.5261 & 0.7001 &  \\
Multi-shot aggregated DMD & 0.8496 & 0.9736 & 0.9903 & 0.9604 & 0.9357 && 0.5312 & 0.7066 &  \\
Multi-shot aggregated reward DMD & \textbf{0.8547} & \textbf{0.9752} & \textbf{0.9921} & \textbf{0.9627} & \textbf{0.9408} && \textbf{0.5394} & \textbf{0.7182} &  \\

\bottomrule
\end{tabular}
}
\end{threeparttable}
\label{tab_ab}
\end{table*}

\subsection{Ablation Study}
To validate the effectiveness of each component and strategy, we conduct ablation studies on both the bidirectional model and the DMD training scheme. All variants are evaluated under the same settings as the main experiments. 

\noindent \textit{\textbf{Effect of history cache retrieval.}}
To evaluate the role of the history cache retrieval, we compare the geometry-based model with and without the history cache. As shown in Tab. \ref{tab_ab}, incorporating the history cache leads to further performance improvement, indicating that accessing historical scene states helps preserve previously generated structures and maintain long-horizon scene consistency during interactive generation. Visual results in Fig. \ref{Fig_ab_1} further show that the history cache improves viewpoint consistency during scene transitions.

\noindent \textit{\textbf{Effect of flow-guided motion preservation.}}
To examine the impact of the flow-guided motion preservation constraint, we compare models trained with and without the proposed motion supervision, as shown in Tab. \ref{tab_ab} and Fig. \ref{Fig_ab_1}. The results show clear improvements in subject consistency, indicating that enforcing temporal coherence in dynamic regions effectively prevents motion drift and promotes more realistic motion of dynamic subjects.

\noindent \textit{\textbf{Effect of the base model architecture.}}
To evaluate the impact of different architectures, we compare the base bidirectional model with a geometry-based model. As shown in Tab. \ref{tab_ab}, the geometry-based model achieves better performance, indicating that geometric priors help stabilize scene structure. In particular, geometric guidance provides explicit spatial cues for scene evolution, thereby reducing structural inconsistencies during viewpoint changes.

\noindent \textbf{\textit{Effect of DMD training scheme.}}
To investigate the impact of different DMD optimization schemes, we compare three variants: (1) Multi-shot DMD, performs gradient updates immediately after each interaction step, (2) Multi-shot aggregated DMD, aggregates the DMD losses over $M$ interaction steps and updates the generator, (3) Multi-shot aggregated reward DMD, further incorporates reward weighting into the aggregated DMD objective. 
As shown in Tab. \ref{tab_ab} and Fig. \ref{Fig_ab_2}, multi-shot aggregated DMD consistently improves performance compared with the step-wise optimization scheme. Further incorporating reward guidance achieves the best performance across all metrics, with clear improvements in visual quality and motion quality.
These results suggest that using an aggregated DMD loss during interactive training is more effective than immediate gradient updates after each step, as it enables the generator to perceive the quality of later interaction outcomes. When combined with reward guidance, the training process is further encouraged to favor higher-quality trajectories, thereby effectively mitigating error accumulation during long-horizon interactions.

\section{Conclusion}
We present MagicWorld, an autoregressive interactive video world model designed to address motion drift of dynamic subjects and error accumulation in long-horizon interactions.
To mitigate motion degradation, we introduce a flow-guided motion preservation constraint enforcing temporal coherence in dynamic regions. To reduce error accumulation, we design a history cache retrieval strategy with an enhanced interactive training scheme.
We also construct RealWM120K, a real-world multimodal dataset for interactive video world modeling.
Extensive experiments show that MagicWorld consistently outperforms existing methods in both VBench metrics and visual quality.

\section{Acknowledgement}
We thank Siming Zheng and Shuolin Xu for their initial support and suggestions to construct  our basic framework.  

{
    \small
    \bibliographystyle{ieeenat_fullname}
    \bibliography{main}
}


\end{document}